\documentclass{article} 
\usepackage[preprint]{colm2026_conference}
\usepackage{microtype}
\usepackage{hyperref}
\usepackage{url}
\usepackage{booktabs}
\usepackage{enumitem}
\usepackage{amsmath}
\usepackage{amssymb}
\usepackage{amsfonts}
\usepackage{graphicx}
\usepackage{subcaption}
\usepackage{multirow}
\usepackage{xspace}

\usepackage{lineno}
\usepackage{pifont}
\usepackage{xcolor}

\usepackage{colortbl}
\usepackage{xcolor}
\definecolor{lightgray}{rgb}{0.9, 0.9, 0.9}
\newcommand{\cmark}{\textcolor{teal}{\ding{51}}}
\newcommand{\xmark}{\textcolor{red!60!black}{\ding{55}}}

\definecolor{darkblue}{rgb}{0, 0, 0.5}
\hypersetup{colorlinks=true, citecolor=darkblue, linkcolor=darkblue, urlcolor=darkblue}

\newcommand{\methodname}{SMETA-ZSL\xspace}

\title{\methodname: Semantic Meta-Alignment for Zero-Shot Threat Classification}

\author{Ivan Alejandro Montoya Sanchez, Anantaa Kotal, Aritran Piplai \\
The University of Texas at El Paso \\
500 W. University Avenue \\
El Paso, TX 79968, USA \\
\texttt{iamontoyasa@miners.utep.edu, \{akotal,apiplai\}@utep.edu}
}

\begin{document}

\ifcolmsubmission
\linenumbers
\fi

\maketitle

\setlength{\belowdisplayskip}{1pt} \setlength{\belowdisplayshortskip}{1pt}
\setlength{\abovedisplayskip}{1pt} \setlength{\abovedisplayshortskip}{1pt}

\begin{abstract}
Cybersecurity systems must adapt rapidly to emerging threats. However, labeled data for new threat categories is unavailable when those threats first appear. Generalized zero-shot learning offers a natural solution by enabling recognition of unseen classes through auxiliary semantic knowledge rather than labeled examples. Large language models are particularly promising in this setting because they can convert unstructured CTI reports into semantic prototypes for emerging threats. However, applying language-driven zero-shot learning to cybersecurity is difficult due to strong semantic overlap between threat descriptions, heterogeneity between behavioral attributes and text, severe class imbalance, and open-set conditions where unseen threats are unknown during training. We propose \methodname, that learns semantic prototypes from overlapping language descriptions through contrastive finetuning, aligns behavioral features through episodic meta-learning and knowledge distillation, and performs adaptive routing for generalization across seen-unseen classes. Across 7 benchmarks, \methodname\ delivers the strongest overall generalized zero-shot performance under the strictest inductive setting, surpassing prior methods by 10.8 points on average, with gains up to 18.1 points. Github: \url{https://github.com/Security-And-Intelligence-Lab-UTEP/SMETA-ZSL}
\end{abstract}

\section{Introduction}
\label{sec:intro}

Cyber defense systems rely on large amounts of raw data, such as malware samples or logs, to train and update models, yet this data is often difficult to obtain, slow to analyze, and not always available when new threats emerge. Consider the task of malware classification: the continuous emergence of new variants and families \citep{tang2023demystifying} necessitates frequent retraining of classifiers, imposing significant computational overhead \citep{li2024revisiting} and making it difficult to obtain timely, labeled data for adapting models to new threats \citep{barros2022malware, aurna2025feedback}. In contrast, Cyber Threat Intelligence (CTI) reports are regularly published by analysts and provide timely natural language descriptions of attacker behavior. However, most defense systems do not operate over natural language, and therefore cannot directly use CTI to adapt or update their behavior, leaving valuable intelligence underutilized. Large Language Models (LLMs) offer a way to process CTI and extract information that can be used by downstream systems. Prior work \citep{chakraborty2026cti, mitra2025falcon, bertiger2025evaluating, schwartz2025llmcloudhunter} has largely focused on using LLMs to generate rule-based defenses. In contrast, leveraging CTI to update data-driven, black-box machine learning systems is significantly more difficult, as it requires converting textual descriptions into signals that can influence model behavior without access to raw data. This gap creates an important language modeling problem: \textit{can language models convert unstructured threat reports into representations that are useful for updating non-linguistic machine learning systems, even when no labeled examples of the new threat exist?} Directly using LLMs for zero-shot malware or attack detection is challenging because cybersecurity artifacts are often too large to analyze in full, making inference costly and frequently exceeding practical context-window limits \citep{qian2025lamd}. Truncating these artifacts into abstract features degrades performance \citep{zhou2024out}. Instead, the goal is to extract and transfer knowledge from CTI using language models to adapt existing cyber-defense systems.

\begin{figure*}[h]
    \centering
    \includegraphics[width=\linewidth]{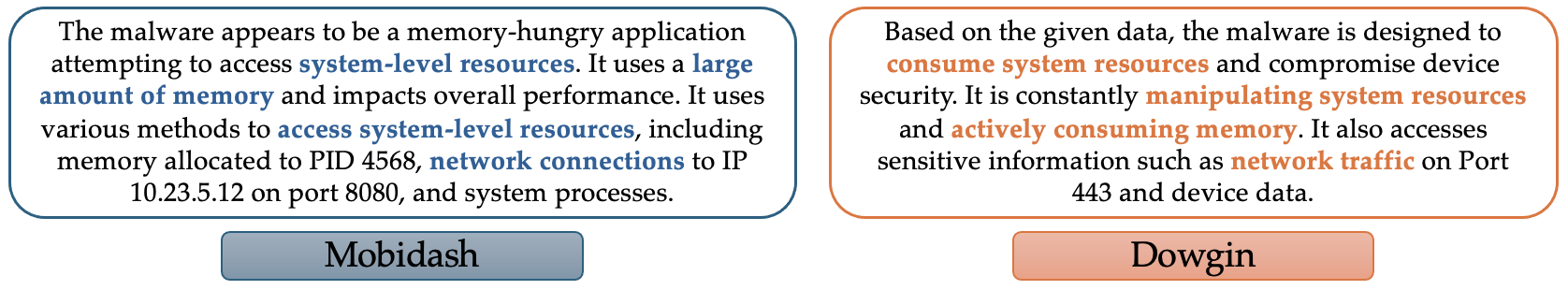}
    \caption{CTI reports for two distinct malware families: Mobidash and Dowgin, share substantial lexical overlap in behavioral descriptions, illustrating why cybersecurity class prototypes are weakly separable in semantic space.}
    \label{fig:cti_overlap}
\end{figure*}

We formulate this problem as a Generalized Zero-Shot Learning (GZSL) setting, where a downstream machine learning model must recognize both seen and unseen classes, with Cyber Threat Intelligence (CTI) providing semantic information for unseen threats. While GZSL with semantic auxiliary information has been extensively explored in generalized settings, such as the vision-language domain \citep{chen2023duet,ma2020variational,rao2024dual, lei2024seeing}, its application to cybersecurity remains an open and largely unsolved challenge, despite being of critical practical importance. The unique challenges in this setup are as follows.
\vspace{-5pt}
\begin{enumerate}[leftmargin=*, noitemsep, partopsep=0pt,topsep=0pt,parsep=0pt]
    \item \textbf{Semantic ambiguity of class prototypes:} Unlike generalized domains with distinguishable semantic features, cybersecurity relies on CTI reports whose descriptions often overlap heavily across malware families. Different families frequently share APIs, behavioral patterns, and attack techniques, making semantic prototypes weakly separable and reducing transfer to unseen classes. Figure~\ref{fig:cti_overlap} illustrates this challenge. 
    
    \item \textbf{Cross-modal heterogeneity:} Malware features are derived from behavioral observations such as API calls, system traces, and network traffic \cite{}. These features differ substantially from the sparse, high-level semantics of CTI reports, making alignment between feature and semantic spaces difficult.
    
    \item \textbf{Class imbalance:} Seen class dominance is already a major challenge in GZSL \cite{}. In cybersecurity, this is amplified by severe imbalance, where benign samples dominate and emerging malware families are rare \cite{}, biasing predictions toward frequent seen classes.
    
    \item \textbf{Open-set classification:} In practice, the number and identity of unseen malware families are unknown during training. Models must therefore distinguish between seen and unseen threats without assuming a fixed set of candidate unseen classes in advance.
\end{enumerate}
\vspace{-5pt}
We propose \textbf{\methodname}, a semantically grounded meta-GZSL framework designed for the realistic cyber threat setting in which unseen classes are not predefined during training, unlabeled unseen instances are unavailable, and adaptation must rely solely on natural-language CTI reports rather than raw artifacts. Unlike prior methods that assume closed sets, access to raw unseen-class data, or predefined unseen prototypes, \methodname operates under a stricter \textit{open-set, class-inductive, instance-inductive setting}. To address the resulting challenges, our framework combines: (1) a contrastively fine-tuned LLM encoder that produces discriminative semantic prototypes from overlapping CTI descriptions; (2) a cross-modal alignment framework that uses episodic meta-learning to explicitly simulate unseen-class emergence during training; and (3) a parameter-free gating mechanism for adaptive inference. Across 7 benchmark datasets, \methodname\ delivers the strongest overall generalized zero-shot performance under the strictest inductive setting.

\section{Background}
\label{sec:background}
ZSL addresses the problem of recognizing instances from classes never observed during training \citep{pourpanah2022review, wang2019survey}. In GZSL, the classifier must operate over the union of both seen and unseen classes simultaneously \citep{li2024improving, verma2020meta}, compounding the difficulty with seen-class dominance — where the posterior probability of seen classes systematically overwhelms that of unseen classes at inference time \citep{kwon2022gating, bhat2025pc, xian2017zero}. Since unseen classes are never observed during training, ZSL methods rely on auxiliary semantic information to bridge the gap. Drawing inspiration from human cognition, where unfamiliar concepts are recognized through background knowledge relating them to familiar ones \citep{fu2015zero, romera2015embarrassingly, tang2024data, li2024improving, sanchez2025semantic}, existing approaches broadly fall into three families: learning a joint embedding space between modalities and semantic representations \citep{ali2023clip, nawaz2022semantically, chen2023duet, lei2024seeing}, synthetically generating unseen-class data from semantic descriptions \citep{mishra2018generative, ma2020variational, tang2024data, wu2025zero, marszalekzeus}, and learning semantic prototypes \citep{wang2025llm, fu2017zero, wang2021dual, rao2024dual, fan2026clip, gill2026llm}. Among these, the semantic prototype approach is particularly well-suited for settings where class-level descriptions are available but instance-level data for unseen classes is absent. ZSL has also been adapted to domain-specific settings, including cybersecurity, where specialized frameworks address malware detection under data scarcity \citep{barros2022malware, wang2025transductive, aurna2025feedback}.

\section{Methodology}
\begin{figure}[h]
    \centering
    \includegraphics[width=\linewidth]{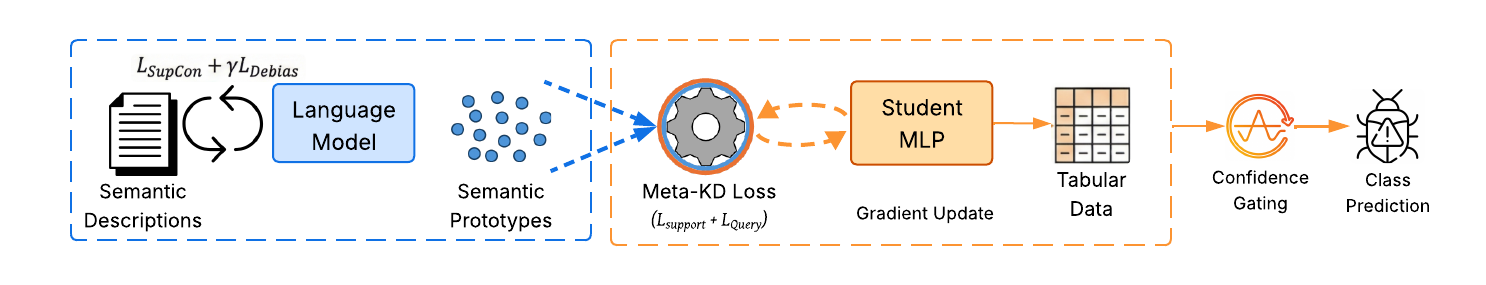}
    \caption{Overview of our proposed \methodname framework that uses semantic prototype for GZSL, when semantic descriptions are overlapping and at a higher level of abstraction.}
    \label{fig:pipeline}
\end{figure}

We propose \methodname, a semantically grounded GZSL framework for tabular data classification. Our core hypothesis is that language models, when appropriately fine-tuned, can serve as a reliable semantic bridge between overlapping natural language descriptions and fine-grained behavioral attributes, enabling zero-shot recognition of unseen classes through episodic meta-training that explicitly aligns heterogeneous modalities. Figure \ref{fig:pipeline} gives an overview of our proposed framework. 

\subsection{Preliminaries}
\label{sec:formulation}
In Zero-shot learning there are two disjoint sets of classes: the seen classes $\mathcal{S}$, for which labeled training instances are available, and the unseen classes $\mathcal{U}$, for which no labeled instances exist at training time. Specifically in GZSL, the classifier $f(\cdot): \mathcal{X} \rightarrow \mathcal{S} \cup \mathcal{U}$ must operate over the union of both seen and unseen classes. Each class $c \in \mathcal{S} \cup \mathcal{U}$ is represented by a prototype vector $t \in \mathcal{T} \subseteq \mathbb{R}^M$, encoded via a prototyping function $\pi(\cdot): \mathcal{S} \cup \mathcal{U} \rightarrow \mathcal{T}$. This yields prototype sets $\mathcal{T}_s = \{t^s_i\}_{i=1}^{N_s}$ and $\mathcal{T}_u = \{t^u_i\}_{i=1}^{N_u}$ for seen and unseen classes respectively, serving as the semantic bridge through which knowledge is transferred from $\mathcal{S}$ to $\mathcal{U}$.  Given a test instance $x_i$, a projection network $f_\theta: \mathcal{X} \rightarrow \mathcal{T}$ maps it into the shared semantic space, and the predicted class is determined by nearest-prototype matching:
\begin{equation}
\hat{y}_i = \arg\max_{c \in \mathcal{S} \cup \mathcal{U}} \frac{f_\theta(x_i) \cdot t_c}{\|f_\theta(x_i)\| \|t_c\|}
\end{equation}

The cybersecurity setting introduces two additional structural constraints that go beyond standard GZSL. First, we operate under the open-set assumption: unlike closed-set GZSL where $\mathcal{S} \cup \mathcal{U}$ is fully predefined, only $\mathcal{S}$ is known at training time. The unseen classes $\mathcal{U}$, including their cardinality $|\mathcal{U}| = N_u$ and identities $\{c^u_i\}_{i=1}^{N_u}$, are revealed only at inference, reflecting the real-world scenario where novel malware families cannot be anticipated in advance. Second, we adopt the Class-Inductive, Instance-Inductive (CIII) setting \citep{pourpanah2022review}, under which the model is trained using only $\mathcal{D}_{tr}$ and $\mathcal{T}_s$, without access to $\mathcal{T}_u$ during training. The unseen prototypes are made available only at inference, serving as the semantic bridge through which the model generalizes to novel unseen classes.  In the cybersecurity domain, this bridge is constructed from cyber threat intelligence (CTI) reports, unstructured expert-authored documents describing behavioral signatures, tactics, and techniques of each threat class. Our goal is to ensure that knowledge transfers reliably from $\mathcal{S}$ to $\mathcal{U}$ despite the semantic overlap and cross-modal heterogeneity these reports introduce, a challenge addressed directly by \methodname.

\subsection{Discriminative Semantic Prototype Learning with LLM}
\label{sec:prototypes}
\textbf{LLM for Semantic Prototyping:} LLMs are natural candidates for semantic prototype construction: given a natural language description of a class, a pretrained encoder can map the text into a dense vector that captures its semantic content. These vectors serve as class prototypes in $\mathcal{T}$, providing a representation of each class  However, naively encoding CTI reports with a pretrained language model yields weakly separable prototypes, as threat descriptions across different classes frequently share structural vocabulary and domain-specific terminology, causing their embeddings to cluster around a shared centroid rather than reflect class-discriminative structure.

\begin{figure}[h]
    \centering
    \begin{subfigure}[t]{0.44\textwidth}
        \centering
        \fbox{\includegraphics[width=\linewidth, height=0.6\textwidth, keepaspectratio=false]{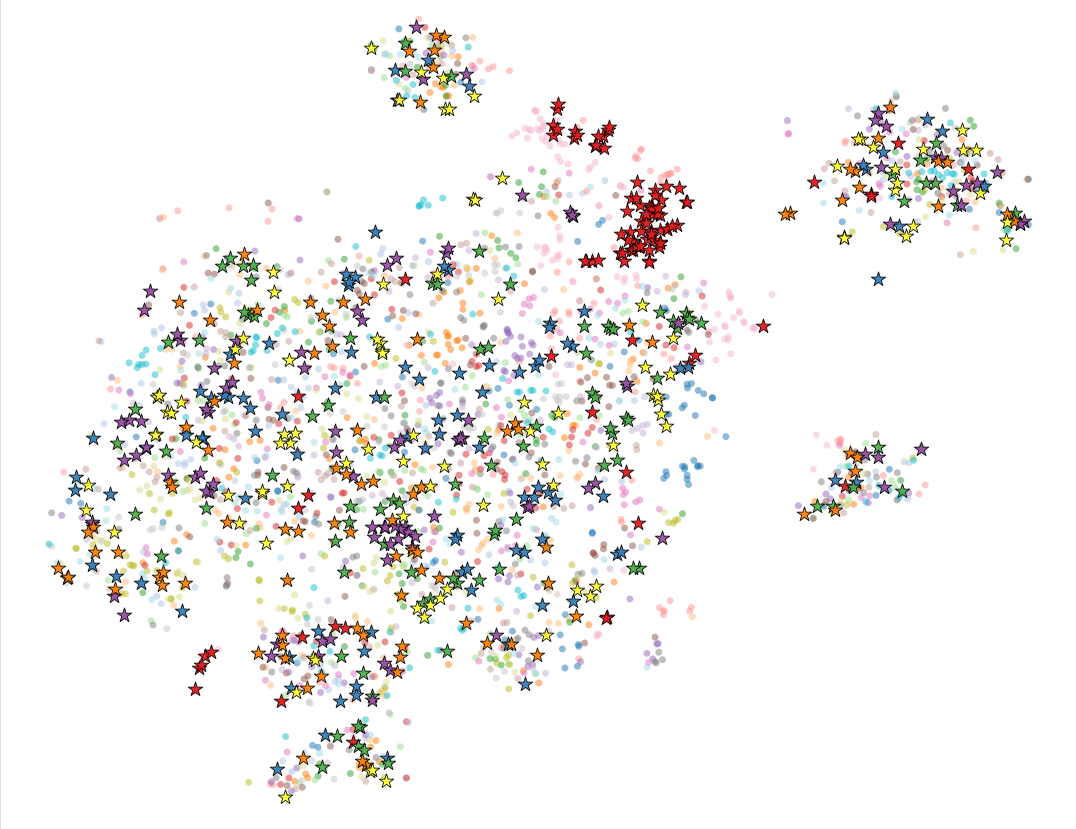}}
        \caption{Semantic embeddings with Pretrained LLM}
        \label{fig:andmal-pre}
    \end{subfigure}\hspace{10pt}
    \begin{subfigure}[t]{0.44\textwidth}
        \centering
        \fbox{\includegraphics[width=\linewidth, height=0.6\textwidth, keepaspectratio=false]{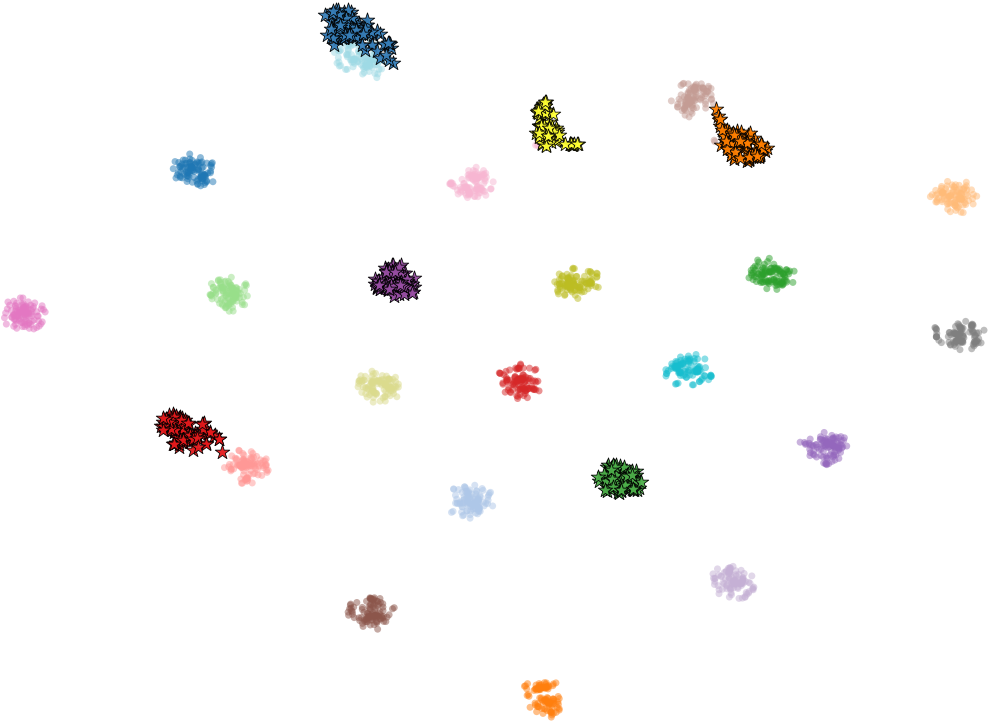}}
        \caption{Semantic embeddings with Contrastively Finetuned LLM}
        \label{fig:andmal-ft}
    \end{subfigure}
    \caption{Comparison of semantic embedding quality with and without contrastive finetuning for datapoints in CIC-AndMal \citep{cicandmal2020} dataset.}
    \label{fig:contrast_finetune}
\end{figure}

\textbf{Supervised Contrastive Objective:} Let $z_i \in \mathbb{R}^M$ denote the L2-normalized embedding of description sample $i$ produced by the language model. To enforce intra-class compactness and inter-class separation in the language model's representation space, we optimize it with a Supervised Contrastive loss \citep{khosla2020supervised}, which pulls descriptions of the same class together while pushing descriptions of different classes apart:
\begin{equation}
\mathcal{L}_{\text{SupCon}} = \sum_{i \in I} \frac{-1}{|P(i)|} \sum_{p \in P(i)} \log \frac{\exp(z_i \cdot z_p / \tau)}{\sum_{a \in A(i)} \exp(z_i \cdot z_a / \tau)}
\end{equation}
where $P(i)$ is the set of positive samples whose descriptions share the same class label as $i$, $A(i)$ is the set of all other description samples in the batch, and $\tau$ is a temperature hyperparameter. Figure~\ref{fig:contrast_finetune} illustrates the effect of contrastive finetuning on the geometry of the learned semantic space.

\textbf{Isotropy Regularization:} Optimizing solely on $\mathcal{L}_{\text{SupCon}}$ is insufficient when class descriptions share substantial structural overlap. The encoder tends to learn the general structure of descriptions rather than class-discriminative signatures, causing embeddings to collapse toward a shared mean. To explicitly counteract this, we introduce an isotropy regularization term that penalizes the mean squared cosine similarity of each embedding $z_i$ to the L2-normalized batch mean $\bar{z}$, discouraging the semantic space from collapsing toward a centroid driven by shared boilerplate vocabulary:
\begin{equation}
\mathcal{L}_{\text{Iso}} = \frac{1}{N} \sum_{i=1}^{N} (z_i \cdot \bar{z})^2
\end{equation}
The combined training objective for the semantic encoder is $\mathcal{L}_{\text{sem}} = \mathcal{L}_{\text{SupCon}} + \gamma \mathcal{L}_{\text{Iso}}$, where $\gamma \geq 0$ controls the strength of the isotropy regularization relative to the contrastive objective.

\textbf{Prototype Construction:} Once the encoder is trained, a class prototype $t_c \in \mathcal{T}$ is computed for each class $c \in \mathcal{S} \cup \mathcal{U}$ as the L2-normalized mean of its constituent embeddings, where $\mathcal{S}_c$ denotes the set of description samples belonging to class $c$. This yields the prototype sets $\mathcal{T}_s$ and $\mathcal{T}_u$, which serve as fixed semantic anchors for all subsequent alignment and inference.

\subsection{Cross-Modal Alignment via Meta Knowledge Distillation}
\label{sec:alignment}
Even with discriminative semantic prototypes, aligning $\mathcal{X}$ with $\mathcal{T}$ remains non-trivial due to the granularity mismatch between precise instance-level behavioral observations and coarse, abstract semantic descriptions. 
We address this through a novel framework that combines meta-knowledge distillation \citep{pan2021meta} with episodic meta-learning, where the zero-shot condition is explicitly simulated at every training step.

\textbf{Two-Tower Architecture:} The framework operates across two representation spaces. The Semantic Tower encodes class descriptions into $\mathcal{T}$, producing the fixed class prototypes $\{t_c\}$ derived in Section~\ref{sec:prototypes}. The Behavioral Tower consists of a projection network $f_\theta: \mathcal{X} \rightarrow \mathcal{T}$ that learns to map behavioral feature vectors into the shared semantic space, producing $\hat{z}_i = f_\theta(x_i) \in \mathbb{R}^M$ for each instance $x_i \in \mathcal{X}$.

\textbf{Episodic Meta-Training:} To prevent $f_\theta$ from overfitting to seen-class alignment and to explicitly enforce generalization to unseen classes, training is structured as a series of meta-learning episodes that simulate the zero-shot condition at every step. At each episode, the set of available training classes $\mathcal{C}_{train}$ is randomly partitioned into a support set $\mathcal{C}_{sup}$ and a query set $\mathcal{C}_{qry}$, where $\mathcal{C}_{train} = \mathcal{C}_{sup} \cup \mathcal{C}_{qry}$ and $\mathcal{C}_{sup} \cap \mathcal{C}_{qry} = \emptyset$. The query set classes are treated as proxy-unseen classes, known training classes deliberately withheld at each episode to simulate novel conditions, forcing $f_\theta$ to generalize its cross-modal alignment beyond the classes it is currently supervised on.

\textbf{Dual-Objective Distillation Loss:} The loss function balances exact knowledge transfer on support classes with generalization on withheld query classes. Let $s_{i,c} = \hat{z}_i \cdot t_c / \|\hat{z}_i\| \|t_c\|$ denote the cosine similarity between the projected instance and prototype $t_c$. For the support set, the student is supervised via a distillation objective that combines soft knowledge transfer through KL divergence against the teacher's similarity distribution with a hard cross-entropy target:
\begin{equation}
\mathcal{L}_{\text{sup}} = \mathcal{L}_{\text{KL}}\left(\sigma\!\left(\frac{s_{i,c}}{T}\right) \,\Big\|\, \sigma\!\left(\frac{\hat{s}_{i,c}}{T}\right)\right) + \mathcal{L}_{\text{CE}}(\hat{z}_i, y_i)
\end{equation}
where $T$ is the distillation temperature, $\hat{s}_{i,c}$ are the teacher's similarity logits, and $\sigma(\cdot)$ denotes the softmax function. The KL divergence transfers the teacher's soft relational structure across classes, while the cross-entropy term anchors the student to the correct hard label. For the query set, no distillation supervision is provided and the student is evaluated purely on prototype matching, acting as a generalization regularizer: $\mathcal{L}_{\text{qry}} = \mathcal{L}_{\text{CE}}(\hat{z}_i, y_i),\ i \in \mathcal{C}_{qry}$. The combined training objective is:
\begin{equation}
\mathcal{L}_{\text{align}} = \mathcal{L}_{\text{sup}} + \lambda \mathcal{L}_{\text{qry}}
\end{equation}
where $\lambda \geq 0$ scales the generalization penalty relative to the distillation objective.

\subsection{Generalized Inference via Adaptive Confidence Gating}
\label{sec:gating}

\begin{figure}[t]
    \centering
    \includegraphics[width=0.8\linewidth]{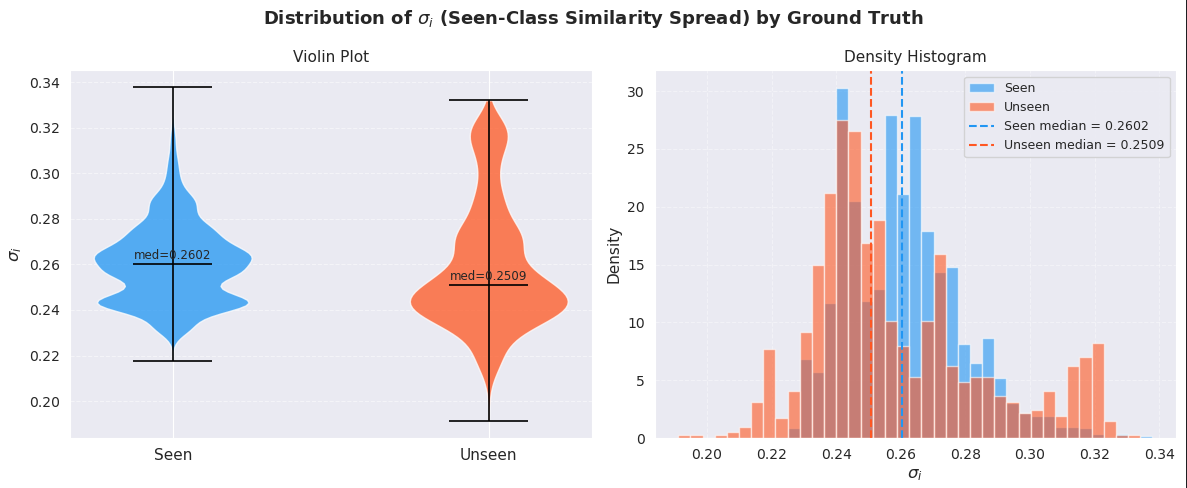}
    \caption{Distribution of per-sample $\sigma_i$ (standard deviation of
    seen-class cosine similarities) stratified by ground truth label.}
    \label{fig:sigma_dist}
\end{figure}

\textbf{Z-Score Confidence Estimation:} The central observation motivating the gating mechanism is that a seen-class instance produces a similarity distribution over seen prototypes with one sharply dominant score, while an unseen-class instance produces a flatter, less discriminative distribution, as empirically confirmed in Figure~\ref{fig:sigma_dist}. Rather than comparing the top seen-class score against a fixed threshold, we evaluate how statistically dominant it is relative to the per-sample distribution of all seen-class similarities. Given a test instance $x_i$, we compute $\hat{z}_i = f_\theta(x_i)$ and the cosine similarity $s_{i,c} = \hat{z}_i \cdot t_c / \|\hat{z}_i\|\|t_c\|$ against every seen prototype $t_c \in \mathcal{T}_s$. The per-sample mean $\mu_i$ and standard deviation $\sigma_i$ of this seen-class similarity distribution are then computed, and the Z-score of the maximum seen-class similarity $s^{\max}_i = \max_{c \in \mathcal{S}} s_{i,c}$ measures how many standard deviations it lies above the mean:
\begin{equation}
Z_i = \frac{s^{\max}_i - \mu_i}{\max(\sigma_i, \epsilon)}
\end{equation}
where $\epsilon$ ensures numerical stability when all seen-class similarities are nearly identical.

\textbf{Routing Decision:} $Z_i$ quantifies how statistically dominant the best seen-class match is for a given sample. A high $Z_i$ indicates that one seen prototype produces a sharply dominant response, suggesting the instance belongs to a seen class. A low $Z_i$ indicates a flat similarity distribution, suggesting the instance is out-of-distribution with respect to $\mathcal{S}$ and likely belongs to an unseen class. The routing decision is formalized as:
\begin{equation}
\hat{y}_i = \begin{cases} \arg\max_{c \in \mathcal{S}}\ s_{i,c} & \text{if } Z_i \geq \tau \\ \arg\max_{c \in \mathcal{U}}\ s_{i,c} & \text{if } Z_i < \tau \end{cases}
\end{equation}
where $\tau$ is a confidence threshold optimized on the validation set to maximize the harmonic mean between seen and unseen class accuracy. This mechanism requires no additional learned parameters and operates entirely on the cosine similarity scores already computed during inference.

\section{Experiments}
\textbf{Baselines:} We evaluate nine methods capable of zero-shot tabular data classification — a critical requirement in cybersecurity — spanning three paradigms: generative \citep{mishra2018generative, verma2020meta}, cybersecurity-specific \citep{wang2025transductive, aurna2025feedback, barros2022malware}, and LLM-based \citep{shi2024surprisingly, wang2025llm, ali2023clip, yun2024tabular}. We additionally evaluate TabPFN \citep{hollmann2023tabpfn}, APT \citep{wu2025apt}, and ZEUS \citep{marszalekzeus}, which generalise to novel datasets with minimal oversight but still require at least one sample from each unseen class at inference time. As such, these models do not strictly satisfy our GZSL formulation and are evaluated in the one-shot setting. Finally, HistGradBoost \citep{pedregosa2011scikit} and XGBoost \citep{chen2016xgboost} serve as standard tabular classification baselines. Table~\ref{tab:baselines} highlights four axes along which methods differ from \methodname. ZET-LLM \citep{shi2024surprisingly}, ProtoLLM \citep{wang2025llm}, and SMELL \citep{barros2022malware} require architectural adaptation for zero-shot inference, while ZEUS, TabPFN, and APT do not support it at all and are therefore included only in the one-shot comparison. Only TabPFN and APT require unseen class prototypes at training time. CLIP-decoder \citep{ali2023clip}, P2T \citep{yun2024tabular}, ZEUS, and TZSL \citep{wang2025transductive} require unlabelled unseen instances during training. Only CLIP-decoder and P2T support open-set recognition among the baselines. \methodname\ is the only method satisfying all four criteria simultaneously. As baselines differ in which assumptions they relax, we evaluate each under its native setting while holding \methodname\ to the strictest assumption throughout.

\begin{table}[h]
\centering
\caption{\small Requirements and capabilities of baseline methods. Checkmarks indicate whether each method supports zero-shot classification natively, trains without unseen prototypes or unlabelled instances, and handles open-set recognition.}
\label{tab:baselines}
\resizebox{\textwidth}{!}{
\begin{tabular}{lllcccc}
\toprule
\textbf{Method} & \textbf{Architecture} & \textbf{Learning Principle} & \shortstack{\textbf{Zero-Shot} \\ \textbf{Classification}} & \shortstack{\textbf{Trains w/o} \\ \textbf{Unseen} \\ \textbf{Prototypes}} & \shortstack{\textbf{Trains w/o} \\ \textbf{Unlabelled Unseen} \\ \textbf{Instances}} & \textbf{Open-Set} \\
\midrule
CVAE-ZSL (\cite{mishra2018generative}) & VAE & Generative, prototype & \cmark & \cmark & \cmark & \xmark \\
MZSL (\cite{verma2020meta}) & GAN + MLP & Generative, meta-learning & \cmark & \cmark & \cmark & \xmark \\
CLIP-decoder (\cite{ali2023clip}) & ViT & Transfer learning & \cmark & \cmark & \xmark & \cmark \\
P2T (\cite{yun2024tabular}) & LLM & Transfer learning & \cmark & \cmark & \xmark & \cmark \\
ZET-LLM (\cite{shi2024surprisingly}) & LLM + MLP & Feature extraction & \cmark (Modified) & \cmark & \cmark & \xmark \\
ProtoLLM (\cite{wang2025llm}) & LLM & Prototype, transfer learning & \cmark (Modified) & \cmark & \cmark & \xmark \\
ZEUS (\cite{marszalekzeus}) & LLM & Federated learning & \xmark & \cmark & \xmark & \xmark \\
TabPFN (\cite{hollmann2023tabpfn}) & LLM & Transductive & \xmark & \xmark & \cmark & \xmark \\
APT (\cite{wu2025apt}) & LLM & Metric learning & \xmark & \xmark & \cmark & \xmark \\
\midrule
\multicolumn{7}{c}{\textit{ZSL for Cybersecurity}} \\
\midrule
SMELL (\cite{barros2022malware}) & MLP & Metric learning & \cmark(Modified) & \cmark & \cmark & \xmark \\
FL-ZSL (\cite{aurna2025feedback}) & MLP & Continual learning & \cmark & \cmark & \cmark & \xmark \\
TZSL (\cite{wang2025transductive}) & VQ-VAE & Transfer learning & \cmark & \cmark & \xmark & \xmark \\
\midrule
\textbf{\methodname} & LLM + MLP & Prototype, meta-learning & \cmark & \cmark & \cmark & \cmark \\
\bottomrule
\end{tabular}
}
\end{table}

\textbf{Datasets:} We evaluate \methodname on seven benchmark datasets spanning four cybersecurity domains CIC-AndMal-2020 (\cite{cicandmal2020}), BODMAS (\cite{yang2021bodmas}), APIGRAPH (\cite{zhang2020enhancing}), AVASTCTU (\cite{bosansky2022avast}) and three general-domain tabular datasets GOODREADS (\cite{wan2018item, wan2019fine}), PETFINDER (\cite{petfinder2018}), FAKEDDIT (\cite{nakamura2020r}). Dataset details including the class distributions are provided in Table~\ref{tab:datasets}.

\begin{table}[h]
\centering
\caption{\small Overview of evaluation datasets spanning cybersecurity and general-domain benchmarks}
\label{tab:datasets}
\resizebox{\textwidth}{!}{%
\begin{tabular}{l l r r r r r}
\toprule
\textbf{Dataset} & \textbf{Features} & \textbf{Size} & \textbf{Semantic Source} & \textbf{Total Classes} & \textbf{Seen} & \textbf{Unseen} \\
\midrule
\rowcolor{lightgray}
CIC-AndMal-2020 & Dynamic behavioral features (API, memory, network, logcat) & 400,000 & CTI reports & 26 & 22 & 4 \\
BODMAS & Static PE features (2,381-dim) & 134,435 & CTI reports & 44 & 40 & 4 \\
\rowcolor{lightgray}
APIGRAPH & Android API calls & 322,594 & CTI reports & 74 & 69 & 5 \\
AVASTCTU & Dynamic sandbox execution logs (CAPEv2) & $\sim$49,000 & CTI reports & 10 & 7 & 3 \\
\rowcolor{lightgray}
GOODREADS & Book metadata (page count, publication year, rating) & 10,000 & Book description & 8 & 6 & 2 \\
PETFINDER & Pet attributes (age, breed, gender) & 15,000 & Adoption profile & 5 & 3 & 2 \\
\rowcolor{lightgray}
FAKEDDIT & Post metadata (score, upvote ratio) & 420,000 & Title/OCR text & 6 & 4 & 2 \\
\bottomrule
\end{tabular}}
\end{table}

\textbf{Implementation Details:} Semantic prototypes are encoded using LLaMA-3.1-8B, loaded in 4-bit precision via \texttt{bitsandbytes} and fine-tuned with LoRA to embed per-class behavioral descriptions into 4,096-dimensional prototype vectors. The projection network $f_\theta$ is a two-block residual MLP with hidden size 1,024, GELU activations, BatchNorm, and dropout ($p{=}0.1$), mapping tabular features into $\mathcal{T}$ with L2-normalized outputs. All hyperparameters are tuned on the validation set; sensitivity analysis is provided in Appendix~\ref{tab:sensitivity}.

\textbf{CTI Report Collection:} CTI reports were collected from two open-source repositories: the ORKL Community CTI Library (\cite{orkl}), a searchable corpus of public threat intelligence reports, and the APT\_REPORT archive (\cite{blackorbird}), a community-maintained collection of vendor and government CTI documents. To ensure semantic coverage across all malware families, particularly for unseen classes with limited report availability, we employed a structured augmentation pipeline: a LLM was prompted to generate synthetic CTI variants conditioned on relevant MITRE ATT\&CK technique descriptors (TTPs) corresponding to each malware family (\cite{attack}). LLM-based generation of CTI has been explored as a viable approach to address report scarcity in threat intelligence workflows (\cite{ranade2021fakecti}).

\textbf{Evaluation Protocol:} For finetuning of the LLM we use a 90/10 split for training and validation. Unseen classes are selected previously and held out during the whole fine-tuning process. During the meta-learning stage, we hold out 15\% of the seen samples as validation and 15\% for test.The number of unseen families varies between 3 and 2 depending on the dataset label size. We follow the standard GZSL evaluation protocol, reporting accuracy on seen classes (S), accuracy on unseen classes (U), and their harmonic mean (H) as the primary metric. The harmonic mean is used as the primary ranking criterion as it penalizes methods that trivially favour either seen or unseen classes. Results are averaged over five random splits and reported with standard deviation.

\section{Results}
\label{sec:results}
\begin{figure}[h]
    \centering
    \includegraphics[width=\linewidth]{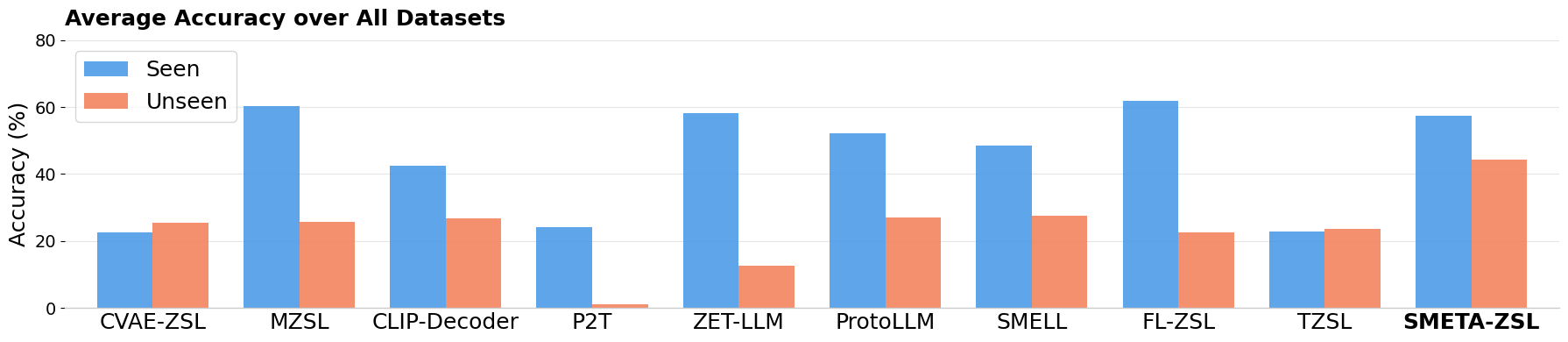}
    \caption{Average seen and unseen class accuracy across all benchmark datasets. Our proposed SMETA-ZSL achieves the highest unseen accuracy among all compared methods while maintaining competitive seen accuracy.}
    \label{fig:seen-unseen-comparison}
\end{figure}

\textbf{GZSL Baseline Comparison:} Table~\ref{tab:gzsl_malware} reports GZSL results across 5 runs for seven tabular benchmarks spanning both cybersecurity and general-domain datasets. On average, \methodname\ improves the harmonic mean by approximately 10.8 points over the strongest baseline on the datasets where it leads, with the largest gains observed on CIC-AndMal (+18.1 over MZSL), GOODREADS (+11.6 over ProtoLLM), AVASTCTU (+12.5 over FL-ZSL) and APIGRAPH (+19.4 over TZSL). On FAKEDDIT, ProtoLLM leads (47.80 vs.\ 44.45), where prototype-based alignment appears particularly well-suited to the dataset's multimodal label structure. \methodname\ is the only method that performs consistently well across all seven datasets under strict inductive assumptions. Figure~\ref{fig:seen-unseen-comparison} further illustrates this trend: \methodname\ attains the highest average unseen accuracy across all seven benchmarks while maintaining competitive seen accuracy. The full results for seen, unseen and mean accuracy across all 5 runs are given in Appendix.

\begin{table}[h]
\renewcommand{\arraystretch}{1.2}
\centering
\caption{GZSL results across seven tabular datasets. \methodname\ is evaluated under the strictest inductive assumptions; baselines under their native settings. Best results (in \textbf{\textcolor{blue}{blue}}) indicate the highest harmonic mean between seen and unseen class accuracy.}
\label{tab:gzsl_malware}
\resizebox{\textwidth}{!}{%
\begin{tabular}{lccccccc}
\toprule
\textbf{Method} & \textbf{CIC-AndMal} & \textbf{BODMAS} & \textbf{APIGRAPH} & \textbf{AVASTCTU} & \textbf{GOODREADS} & \textbf{PETFINDER} & \textbf{FAKEDDIT} \\
\midrule
CVAE-ZSL     & $17.56 \pm 2.61$ & $30.20 \pm 2.50$ & $8.22 \pm 2.32$   & $24.85 \pm 7.23$  & $11.89 \pm 1.72$ & $1.46 \pm 1.95$   & $14.33 \pm 4.05$  \\
MZSL         & $39.70 \pm 5.89$ & $42.23 \pm 4.97$ & $23.21 \pm 2.62$  & $22.27 \pm 14.74$ & $22.50 \pm 0.72$ & $30.96 \pm 1.05$  & $38.16 \pm 2.19$  \\
CLIP         & $14.34 \pm 3.72$ & $43.15 \pm 8.39$ & $21.49 \pm 4.55$  & $39.14 \pm 6.46$  & $18.76 \pm 0.90$ & $32.24 \pm 1.25$  & $36.56 \pm 14.66$ \\
P2T          & $0.00 \pm 0.00$  & $0.00 \pm 0.00$  & $0.47 \pm 0.49$   & $0.48 \pm 0.10$   & $7.81 \pm 0.93$  & $2.52 \pm 0.68$   & $0.55 \pm 0.35$   \\
ZET-LLM      & $13.03 \pm 1.20$ & $32.32 \pm 8.36$ & $12.67 \pm 1.57$  & $0.01 \pm 0.01$   & $20.48 \pm 0.55$ & $23.14 \pm 0.78$  & $1.05 \pm 0.19$   \\
ProtoLLM     & $38.15 \pm 0.67$ & $37.01 \pm 0.19$ & $21.59 \pm 0.16$  & $42.26 \pm 4.15$  & $24.34 \pm 0.25$ & $25.69 \pm 0.91$  & $\mathbf{\textcolor{blue}{47.80 \pm 0.20}}$  \\
SMELL        & $28.35 \pm 1.48$ & $27.60 \pm 2.18$ & $12.32\pm 5.45$  & $36.71 \pm 1.09$  & $24.21 \pm 1.54$ & $3.15 \pm 0.59$   & $21.10 \pm 5.52$  \\
FL-ZSL       & $29.43 \pm 3.27$ & $48.12 \pm 1.76$ & $27.16 \pm 2.53$  & $44.46 \pm 5.54$  & $21.89 \pm 1.02$ & $31.41 \pm 0.42$  & $21.20 \pm 4.65$  \\
TZSL         & $10.94 \pm 4.88$ & $41.20 \pm 9.47$ & $30.80 \pm 3.59$  & $29.67 \pm 23.09$ & $0.00 \pm 0.00$  & $6.13 \pm 0.37$   & $17.58 \pm 1.78$  \\
\midrule
\textbf{\methodname} & $\mathbf{\textcolor{blue}{57.78 \pm 1.02}}$ & $\mathbf{\textcolor{blue}{50.20 \pm 9.10}}$ & $\mathbf{\textcolor{blue}{50.19 \pm 3.61}}$ & $\mathbf{\textcolor{blue}{57.00 \pm 1.22}}$ & $\mathbf{\textcolor{blue}{35.92 \pm 2.50}}$ & $\mathbf{\textcolor{blue}{33.36 \pm 0.58}}$ & $44.45\pm 4.28$ \\
\bottomrule
\end{tabular}}
\end{table}

\textbf{One-Shot Baseline Comparison:} Table~\ref{tab:oneshot_comparison} compares \methodname\ (zero-shot) against few-shot and standard tabular baselines. \methodname\ achieves the best harmonic mean on five of seven benchmarks, with margins of +15.0 on GOODREADS, +10.4 on PETFINDER, and +7.1 on APIGRAPH. On BODMAS and AVASTCTU, HistGradBoost and ZEUS respectively benefit from labelled support examples unavailable to \methodname. Despite strictly zero-shot assumptions, \methodname\ remains the most consistently strong method across the suite. Fully supervised upper-bound results are in Appendix~\ref{tab:supervised_ceiling}.

\begin{table}[h]
\renewcommand{\arraystretch}{1.2}
\centering
\caption{One-shot classification results across seven tabular benchmarks. \methodname\ is compared against methods designed for few-shot tabular generalisation and standard tabular classifiers. Best results in \textbf{\textcolor{blue}{blue}}.}
\label{tab:oneshot_comparison}
\resizebox{\textwidth}{!}{%
\begin{tabular}{l ccccccc}
\toprule
\textbf{Method} & \textbf{CIC-AndMal} & \textbf{BODMAS} & \textbf{APIGRAPH} & \textbf{AVASTCTU} & \textbf{GOODREADS} & \textbf{PETFINDER} & \textbf{FAKEDDIT} \\
\midrule
TabPFN         & $36.28 \pm 0.38$ & $49.09 \pm 5.41$ & $4.91 \pm 6.61$  & $42.14 \pm 1.31$ & $0.00 \pm 0.00$ & $0.00 \pm 0.00$ & $12.66 \pm 8.36$ \\
APT            & $51.92 \pm 1.10$ & $49.44 \pm 0.72$ & $11.54 \pm 0.64$ & $65.52 \pm 0.66$ & $20.57 \pm 0.74$ & $14.82 \pm 1.40$ & $10.59 \pm 3.07$ \\
ZEUS           & $47.29 \pm 0.69$ & $58.72 \pm 4.22$ & $43.12 \pm 0.93$ & $\mathbf{\textcolor{blue}{79.87 \pm 3.31}}$ & $20.97 \pm 3.77$ & $23.00 \pm 1.97$ & $40.84 \pm 3.50$ \\
HistGradBoost  & $37.21 \pm 1.09$ & $\mathbf{\textcolor{blue}{59.83 \pm 3.18}}$ & $38.93 \pm 0.96$ & $61.70 \pm 2.11$ & $0.02 \pm 0.02$  & $0.57 \pm 0.06$  & $1.04 \pm 0.15$  \\
XGBoost        & $24.57 \pm 7.60$ & $27.85 \pm 17.82$ & $1.67 \pm 1.08$ & $9.26 \pm 10.32$ & $0.00 \pm 0.00$ & $0.00 \pm 0.00$  & $0.00 \pm 0.00$  \\
\midrule
\textbf{\methodname} & $\mathbf{\textcolor{blue}{56.7 \pm 1.6}}$ & $50.19 \pm 3.61$ & $\mathbf{\textcolor{blue}{50.20 \pm 9.10}}$ & $57.00 \pm 1.22$ & $\mathbf{\textcolor{blue}{35.92 \pm 2.50}}$ & $\mathbf{\textcolor{blue}{33.36 \pm 0.58}}$ & $\mathbf{\textcolor{blue}{44.45\pm 4.28}}$  \\
\bottomrule
\end{tabular}}
\end{table}

\textbf{Effect of Embedding Model:} Table~\ref{tab:embedding} reports the effect of substituting the semantic encoder on GZSL performance, holding the remainder of \methodname fixed. LLaMA3.1-8B is the best-performing encoder overall, achieving the highest scores on both CIC-AndMAL (56.70 ± 1.6) and BODMAS (50.19 ± 3.61). It also exhibits relatively low variance, suggesting stable semantic prototypes across splits. 

\begin{table}[h]
\centering
\caption{Effect of language model on \methodname}
\label{tab:embedding}
\resizebox{\textwidth}{!}{%
\begin{tabular}{l cccccc}
\toprule
\textbf{Dataset} & \textbf{Qwen-3-4B} & \textbf{Gemma-3-4B} & \textbf{LLaMA3.1-3B} & \textbf{LLaMA3.1-8B} & \textbf{Mistral-Nemo-Base-12B} & \textbf{Qwen-14B} \\
\midrule
CIC-AndMAL & $55.45 \pm 2.03$ & $54.34 \pm 3.57$ & $45.61 \pm 4.93$ & $56.70 \pm 1.6$ & $37.14 \pm 6.79$ & $48.58 \pm 3.36$ \\
BODMAS     & $33.25 \pm 2.76$ & $48.71 \pm 2.29$ & $41.68 \pm 1.96$ & \textbf{$50.19 \pm 3.61$} & $33.25 \pm 2.76$ & $48.30 \pm 5.02$ \\
\bottomrule
\end{tabular}}
\end{table}

\textbf{Ablation Studies:} Table~\ref{tab:ablation} reports each component's contribution via systematic ablation. Removing LLM semantic embeddings drops the harmonic mean by 9.3 points on average, confirming prototype quality as the primary performance driver. Replacing episodic meta-learning with standard training degrades unseen-class accuracy substantially ($-5.11$ on APIGRAPH, $-4.88$ on AVASTCTU), validating that simulating the zero-shot condition during training is critical. Removing knowledge distillation yields a more moderate but consistent drop, indicating soft relational transfer provides complementary signal beyond hard label supervision.

\textbf{Few-Shot Setting:} Appendix~\ref{tab:kshot} examines how \methodname\ performs as labeled samples per unseen class increase from zero to five. Performance generally continues to improve with more shots, with gains of up to 18.8 points on AVASTCTU at $K=5$, though with diminishing returns beyond $K=2$.

\begin{table}[h]
\centering
\caption{Ablation study across cybersecurity datasets. Each row removes or replaces one component of \methodname.}
\label{tab:ablation}
\resizebox{\textwidth}{!}{%
\begin{tabular}{l cccc}
\toprule
\textbf{Configuration} & \textbf{CIC-AndMAL} & \textbf{BODMAS} & \textbf{APIGRAPH} & \textbf{AVASTCTU} \\
\midrule
\rowcolor{lightgray}
\textbf{Full \methodname}
    & $\textbf{56.7}$ & \textbf{59.10} & \textbf{50.19} & \textbf{57.00} \\
w/o LLM semantic embeddings (random init)
    & $36.12$ & 48.71  & 46.12 & 54.92 \\
w/o meta-learning episodes
    & $54.63$ & 59.11 & 45.08 & 52.12 \\
w/o knowledge distillation
    & $55.78$ & 57.41 & 44.55 & 52.12 \\
Replace LLM with static GloVe vectors
    & $22.99$ & 33.95 & 39.84 & 49.33 \\
\bottomrule
\end{tabular}}
\end{table}

\textbf{Effect of Synthetic CTI:} Our problem space has a core data scarcity issue:  the number of real CTI reports per malware family is too small to provide sufficient within-class diversity for optimizing the supervised contrastive objective. Large-scale paired malware-CTI data is difficult to obtain in practice, as organizations often release malware artifacts or CTI descriptions, but rarely both~\cite{johnson2016guide,misp2024}. This is precisely the premise of our problem setting: if the malware artifact is available, its corresponding attack description may not be, and conversely, if a CTI report describing an attack is released, the real malware sample may not be shared. 

To verify that training-time synthetic CTI does not confer an unfair advantage at test time, we ablate synthetic CTI entirely for unseen classes (i.e., classes present 
only at inference), constructing unseen-class prototypes from real CTI descriptions only, while keeping seen-class training unchanged. As shown in Table~\ref{tab:synthetic_cti_ablation}, unseen accuracy ($U$) and harmonic mean ($H$) are statistically indistinguishable 
with and without synthetic CTI across both datasets, confirming that inference-time performance is driven by learned cross-modal alignment rather than synthetic CTI influencing prototype construction.

\begin{table}[t]
\centering
\caption{Ablation on synthetic CTI usage for unseen-class prototype construction. 
$S$: seen accuracy, $U$: unseen accuracy, $H$: harmonic mean.}
\label{tab:synthetic_cti_ablation}
\resizebox{0.6\textwidth}{!}{%
\begin{tabular}{llccc}
\toprule
\textbf{Dataset} & \textbf{Unseen Prototype Source} & \textbf{$S$} & \textbf{$U$} & \textbf{$H$} \\
\midrule
\multirow{2}{*}{CIC-AndMal} & Real CTI only        & 53.75 & 50.00 & 51.81 \\
                            & Real + Synthetic CTI & 53.75 & 50.67 & 52.16 \\
\midrule
\multirow{2}{*}{BODMAS}     & Real CTI only        & 66.75 & 35.00 & 45.92 \\
                            & Real + Synthetic CTI & 66.56 & 33.50 & 44.57 \\
\bottomrule
\end{tabular}
}
\end{table}

\textbf{Reproducibility Statement:} All experiments reporting mean ± std are repeated over 5 runs with different seeds. Per seed result reported in Appendix. Hyperparameters are listed in Appendix \ref{tab:sensitivity}. Semantic prototypes were generated with Llama-3.1-8B and are released as fixed .npy files to avoid non-determinism. Training was performed on 3× NVIDIA RTX ADA 48GB GPUs. Code, checkpoints, and preprocessing scripts are available at \url{https://github.com/Security-And-Intelligence-Lab-UTEP/SMETA-ZSL/blob/main/README.md}

\section{Conclusion}
\vspace{-3mm}
We presented \methodname, a framework for a realistic but underexplored cyber-defense setting in which new threat classes emerge without labeled behavioral examples, unlabeled unseen instances, or prior knowledge of unseen classes. In practice, the only available supervision often comes from natural-language CTI reports. We show that language models can act as semantic adaptation interfaces for downstream cyber-defense systems by converting CTI into discriminative semantic prototypes that transfer to structured behavioral data. Across seven benchmarks spanning four cybersecurity and three general-domain datasets, \methodname~ consistently outperforms prior methods under stricter open-set assumptions, improving baseline by an average of 10.8 points.

\bibliography{colm2026_conference}
\bibliographystyle{colm2026_conference}

\appendix
\section{Appendix}

\begin{table}[h]
\centering
\caption{Fully supervised ceiling performance across seven tabular benchmarks. Each method is trained and evaluated on all classes with full label access, representing the upper bound that zero-shot methods aspire to approach. Results are mean $\pm$ standard deviation over five runs.}
\label{tab:supervised_ceiling}
\resizebox{\textwidth}{!}{%
\begin{tabular}{l ccccccc}
\toprule
\textbf{Method} & \textbf{CIC-AndMal-2020} & \textbf{BODMAS} & \textbf{APIGRAPH} & \textbf{AVASTCTU} & \textbf{GOODREADS} & \textbf{PETFINDER} & \textbf{FAKEDDIT} \\
\midrule
TabPFN          & $85.86 \pm 1.05$ & $94.92 \pm 0.33$ & $77.08 \pm 1.15$ & $99.35 \pm 0.30$ & $58.91 \pm 0.66$ & $63.75 \pm 0.62$ & $99.99 \pm 0.02$ \\
APT             & $66.85 \pm 0.81$ & $56.26 \pm 0.53$ & $15.52 \pm 1.09$ & $98.55 \pm 0.21$ & $41.97 \pm 0.39$ & $61.97 \pm 1.19$ & $99.99 \pm 0.02$ \\
ZEUS            & $56.40 \pm 1.90$ & $82.84 \pm 1.36$ & $57.59 \pm 0.69$ & $93.26 \pm 1.07$ & $32.81 \pm 2.28$ & $38.64 \pm 2.26$ & $91.16 \pm 1.08$ \\
HistGradBoost   & $83.76 \pm 1.44$ & $93.62 \pm 0.81$ & $78.83 \pm 1.27$ & $98.78 \pm 0.16$ & $58.69 \pm 0.86$ & $36.14 \pm 0.45$ & $99.99 \pm 0.01$ \\
XGBoost         & $83.27 \pm 1.70$ & $94.06 \pm 0.57$ & $79.63 \pm 1.25$ & $98.81 \pm 0.15$ & $59.33 \pm 0.67$ & $39.53 \pm 0.94$ & $99.99 \pm 0.01$ \\
MLP    & $77.15 \pm 0.15$ & $90.72 \pm 0.59$ & $77.88 \pm 1.03$ & $98.67 \pm 0.15$ & $47.71 \pm 1.28$ & $36.08 \pm 0.97$ & $98.91 \pm 0.68$ \\
\bottomrule
\end{tabular}}
\end{table}

\begin{table}[h]
\centering
\caption{Sensitivity of SMETA to key hyperparameters across cybersecurity datasets. $H$ = Harmonic Mean (\%). One parameter is varied at a time while others are fixed to their default values (marked with $\star$). Stable $H$ across values indicates low sensitivity to that parameter.}
\label{tab:sensitivity}
\resizebox{0.65\columnwidth}{!}{%
\begin{tabular}{llccc}
\toprule
\textbf{Hyperparameter} & \textbf{Value} & \textbf{CIC-AndMAL} & \textbf{BODMAS} & \textbf{APIGRAPH} \\
\midrule
\multirow{6}{*}{Query Loss $\lambda_{qry}$} 
    & 0.1          & 53.7 & 64.1 & 33.3 \\
    & 0.5          & 50.6 & 59.2 & 29.3 \\
    & 1.0          & 46.7 & 50.6 & 33.7 \\
    & 2.0$^\star$  & 44.8 & 47.9 & 32.6 \\
    & 3.0          & 42.9 & 47.8 & 33.2 \\
    & 4.0          & 42.4 & 47.8 & 31.7 \\
\midrule
\multirow{5}{*}{KD Temp $t_{kd}$} 
    & 1.0          & 44.6 & 48.5 & 32.6 \\
    & 2.0          & 45.0 & 48.2 & 32.6 \\
    & 3.0$^\star$  & 44.8 & 47.9 & 32.6 \\
    & 4.0          & 44.8 & 48.2 & 32.4 \\
    & 5.0          & 44.5 & 48.0 & 32.4 \\
\midrule
\multirow{5}{*}{Alpha $\alpha$} 
    & 0.1          & 48.3 & 53.6 & 33.6 \\
    & 0.3          & 46.8 & 51.0 & 33.4 \\
    & 0.5          & 45.9 & 50.1 & 32.3 \\
    & 0.7$^\star$  & 44.8 & 47.9 & 32.6 \\
    & 0.9          & 42.5 & 51.3 & 32.4 \\
\midrule
\multirow{5}{*}{Ratio (S, U)} 
    & (15, 3)      & 35.9 & 36.9 & 33.9 \\
    & (13, 5)$^\star$ & 44.8 & 47.9 & 32.6 \\
    & (10, 8)      & 51.7 & 46.9 & 35.2 \\
    & (8, 10)      & 53.0 & 61.0 & 30.5 \\
    & (5, 13)      & 54.4 & 62.4 & 33.4 \\
\bottomrule
\end{tabular}%
}
\end{table}

\vspace{-3mm}

\begin{table}[h]
\centering
\caption{GZSL results on CIC-AndMal-2020. Seen = Seen Acc (\%), Unseen = Unseen Acc (\%), Mean = H-Mean (\%). Best Mean in \textbf{bold}.}
\label{tab:gzsl_cicandmal}
\resizebox{\textwidth}{!}{%
\begin{tabular}{l cccccccccc}
\toprule
\textbf{Class} & \textbf{CVAE-ZSL} & \textbf{MZSL} & \textbf{CLIP-Decoder} & \textbf{P2T} & \textbf{ZET-LLM} & \textbf{ProtoLLM} & \textbf{SMELL} & \textbf{FL-ZSL} & \textbf{TZSL} & \textbf{\methodname} \\
\midrule
\multicolumn{11}{c}{\textit{Run 1 (Seed 42)}} \\
\midrule
Seen   & 48.23 & 57.76 & 24.59 & 5.41 & 56.22 & 47.87 & 19.89 & 61.88 & 11.29 & 66.59 \\
Unseen & 9.10  & 27.73 & 6.50  & 0.00 & 6.54  & 33.05 & 56.26 & 22.97 & 25.87 & 49.54 \\
Mean   & 15.32 & 37.47 & 10.28 & 0.00 & 11.72 & 39.10 & 29.39 & 33.50 & 15.72 & \textbf{56.81} \\
\midrule
\multicolumn{11}{c}{\textit{Run 2 (Seed 123)}} \\
\midrule
Seen   & 48.69 & 60.12 & 18.59 & 5.41 & 56.25 & 43.97 & 21.05 & 61.29 & 8.00  & 66.12 \\
Unseen & 11.06 & 33.29 & 21.69 & 0.00 & 8.40  & 32.83 & 55.90 & 20.53 & 17.40 & 53.71 \\
Mean   & 18.02 & 42.86 & 20.02 & 0.00 & 14.62 & 37.59 & 30.58 & 30.76 & 10.96 & \textbf{59.27} \\
\midrule
\multicolumn{11}{c}{\textit{Run 3 (Seed 456)}} \\
\midrule
Seen   & 47.18 & 59.06 & 18.82 & 5.29 & 56.13 & 45.91 & 18.81 & 62.71 & 10.00 & 66.94 \\
Unseen & 8.57  & 42.81 & 9.40  & 0.00 & 6.92  & 33.27 & 56.17 & 17.05 & 6.03  & 51.62 \\
Mean   & 14.51 & 49.64 & 12.54 & 0.00 & 12.32 & 38.58 & 28.19 & 26.81 & 7.53  & \textbf{58.29} \\
\midrule
\multicolumn{11}{c}{\textit{Run 4 (Seed 789)}} \\
\midrule
Seen   & 49.14 & 70.47 & 19.29 & 5.42 & 56.57 & 44.65 & 17.66 & 61.76 & 11.65 & 66.12 \\
Unseen & 11.99 & 23.20 & 13.23 & 0.00 & 7.94  & 32.40 & 55.10 & 16.01 & 24.13 & 50.93 \\
Mean   & 19.28 & 34.91 & 15.69 & 0.00 & 13.93 & 37.55 & 26.74 & 25.43 & 15.71 & \textbf{57.54} \\
\midrule
\multicolumn{11}{c}{\textit{Run 5 (Seed 2025)}} \\
\midrule
Seen   & 48.61 & 58.94 & 18.47 & 5.41 & 56.07 & 44.52 & 17.35 & 60.47 & 2.71  & 67.06 \\
Unseen & 13.13 & 23.55 & 10.21 & 0.00 & 7.09  & 33.04 & 59.12 & 20.53 & 20.88 & 49.54 \\
Mean   & 20.68 & 33.65 & 13.15 & 0.00 & 12.58 & 37.93 & 26.82 & 30.66 & 4.79  & \textbf{56.98} \\
\midrule
\multicolumn{11}{c}{\textit{Average $\pm$ Std}} \\
\midrule
Seen   & $48.37_{\pm 0.74}$ & $61.27_{\pm 4.66}$ & $19.95_{\pm 2.61}$ & $5.39_{\pm 0.05}$  & $56.25_{\pm 0.19}$ & $45.38_{\pm 1.56}$ & $18.95_{\pm 1.38}$ & $61.62_{\pm 0.82}$ & $8.73_{\pm 3.66}$ & $\mathbf{66.56_{\pm 0.44}}$ \\
Unseen & $10.77_{\pm 1.92}$ & $30.12_{\pm 7.32}$ & $12.20_{\pm 5.82}$ & $0.00_{\pm 0.00}$  & $7.38_{\pm 0.77}$  & $32.92_{\pm 0.33}$ & $56.51_{\pm 1.37}$ & $19.42_{\pm 2.84}$ & $18.86_{\pm 7.87}$ & $\mathbf{51.07_{\pm 1.73}}$ \\
Mean   & $17.56_{\pm 2.61}$ & $39.70_{\pm 5.89}$ & $14.34_{\pm 3.72}$ & $0.00_{\pm 0.00}$  & $13.03_{\pm 1.20}$ & $38.15_{\pm 0.67}$ & $28.35_{\pm 1.48}$ & $29.43_{\pm 3.27}$ & $10.94_{\pm 4.88}$ & $\mathbf{57.78_{\pm 1.02}}$ \\
\bottomrule
\end{tabular}}
\end{table}

\begin{table}[h]
\centering
\caption{GZSL results on BODMAS. Seen = Seen Acc (\%), Unseen = Unseen Acc (\%), Mean = H-Mean (\%). Best Mean in \textbf{bold}.}
\label{tab:gzsl_bodmas}
\resizebox{\textwidth}{!}{%
\begin{tabular}{l cccccccccc}
\toprule
\textbf{Class} & \textbf{CVAE-ZSL} & \textbf{MZSL} & \textbf{CLIP-Decoder} & \textbf{P2T} & \textbf{ZET-LLM} & \textbf{ProtoLLM} & \textbf{SMELL} & \textbf{FL-ZSL} & \textbf{TZSL} & \textbf{\methodname} \\
\midrule
\multicolumn{11}{c}{\textit{Run 1 (Seed 42)}} \\
\midrule
Seen   & 30.50 & 92.38 & 62.44 & 3.31 & 85.31 & 72.64 & 21.46 & 64.69 & 57.56 & 83.25 \\
Unseen & 37.17 & 24.17 & 33.50 & 0.00 & 18.67 & 25.00 & 27.00 & 41.50 & 58.67 & 50.00 \\
Mean   & 33.50 & 38.31 & 43.60 & 0.00 & 30.63 & 37.20 & 23.92 & 50.56 & 58.11 & \textbf{62.48} \\
\midrule
\multicolumn{11}{c}{\textit{Run 2 (Seed 123)}} \\
\midrule
Seen   & 23.75 & 92.75 & 60.44 & 3.50 & 86.31 & 70.20 & 43.96 & 64.06 & 45.00 & 76.31 \\
Unseen & 32.00 & 24.00 & 19.50 & 0.00 & 28.00 & 24.83 & 21.67 & 37.00 & 32.50 & 32.00 \\
Mean   & 27.26 & 38.13 & 29.49 & 0.00 & 42.28 & 36.69 & 29.03 & 46.91 & 37.74 & \textbf{45.09} \\
\midrule
\multicolumn{11}{c}{\textit{Run 3 (Seed 456)}} \\
\midrule
Seen   & 30.44 & 90.56 & 64.69 & 3.50 & 85.06 & 72.98 & 47.56 & 65.00 & 46.00 & 82.31 \\
Unseen & 32.67 & 30.33 & 34.83 & 0.00 & 12.00 & 24.83 & 22.17 & 36.17 & 31.17 & 44.00 \\
Mean   & 31.51 & 45.45 & 45.28 & 0.00 & 21.03 & 37.06 & 30.24 & 46.47 & 37.16 & \textbf{57.35} \\
\midrule
\multicolumn{11}{c}{\textit{Run 4 (Seed 789)}} \\
\midrule
Seen   & 28.75 & 92.69 & 57.38 & 3.50 & 85.56 & 71.83 & 37.01 & 64.06 & 38.88 & 74.00 \\
Unseen & 32.33 & 24.50 & 48.33 & 0.00 & 17.50 & 25.00 & 20.83 & 40.17 & 34.00 & 31.00 \\
Mean   & 30.44 & 38.76 & \textbf{52.47} & 0.00 & 29.06 & 37.09 & 26.66 & 49.38 & 36.27 & 43.70 \\
\midrule
\multicolumn{11}{c}{\textit{Run 5 (Seed 2025)}} \\
\midrule
Seen   & 25.25 & 91.94 & 63.25 & 3.50 & 86.50 & 71.29 & 32.01 & 63.56 & 42.13 & 74.31 \\
Unseen & 32.17 & 34.83 & 34.83 & 0.00 & 24.83 & 25.00 & 25.17 & 37.67 & 32.50 & 29.67 \\
Mean   & 28.29 & \textbf{50.52} & 44.93 & 0.00 & 38.59 & 37.02 & 28.18 & 47.30 & 36.69 & 42.40 \\
\midrule
\multicolumn{11}{c}{\textit{Average $\pm$ Std}} \\
\midrule
Seen   & $27.74_{\pm 3.08}$ & $92.06_{\pm 0.80}$ & $61.64_{\pm 2.84}$ & $3.46_{\pm 0.08}$  & $85.75_{\pm 0.63}$ & $71.79_{\pm 1.11}$ & $36.40_{\pm 9.21}$ & $64.28_{\pm 0.57}$ & $45.91_{\pm 7.08}$ & $\mathbf{78.04_{\pm 4.43}}$ \\
Unseen & $33.27_{\pm 2.19}$ & $27.57_{\pm 4.34}$ & $34.20_{\pm 10.21}$ & $0.00_{\pm 0.00}$  & $20.20_{\pm 6.31}$ & $24.93_{\pm 0.09}$ & $23.37_{\pm 2.33}$ & $38.50_{\pm 2.25}$ & $37.77_{\pm 11.73}$ & $\mathbf{37.33_{\pm 9.11}}$ \\
Mean   & $30.20_{\pm 2.50}$ & $42.23_{\pm 4.97}$ & $43.15_{\pm 8.39}$ & $0.00_{\pm 0.00}$  & $32.32_{\pm 8.36}$ & $37.01_{\pm 0.19}$ & $27.60_{\pm 2.18}$ & $48.12_{\pm 1.76}$ & $41.20_{\pm 9.47}$ & $\mathbf{50.20_{\pm 9.10}}$ \\
\bottomrule
\end{tabular}}
\end{table}

\begin{table}[h]
\centering
\caption{GZSL results on APIGRAPH. Seen = Seen Acc (\%), Unseen = Unseen Acc (\%), Mean = H-Mean (\%). Best Mean in \textbf{bold}.}
\label{tab:gzsl_apigraph}
\resizebox{\textwidth}{!}{%
\begin{tabular}{l cccccccccc}
\toprule
\textbf{Class} & \textbf{CVAE-ZSL} & \textbf{MZSL} & \textbf{CLIP-Decoder} & \textbf{P2T} & \textbf{ZET-LLM} & \textbf{ProtoLLM} & \textbf{SMELL} & \textbf{FL-ZSL} & \textbf{TZSL} & \textbf{\methodname} \\
\midrule
\multicolumn{11}{c}{\textit{Run 1 (Seed 42)}} \\
\midrule
Seen   & 23.38 & 71.13 & 58.40 & 1.45 & 34.62 & 43.77 & 53.83 & 53.08 & 32.23 & 46.32 \\
Unseen & 7.36  & 12.34 & 10.54 & 0.00 & 6.80  & 14.58 & 4.23  & 19.04 & 44.55 & 57.92 \\
Mean   & 11.20 & 21.02 & 17.86 & 0.00 & 11.37 & 21.87 & 7.85  & 28.03 & 37.40 & \textbf{51.48} \\
\midrule
\multicolumn{11}{c}{\textit{Run 2 (Seed 123)}} \\
\midrule
Seen   & 25.39 & 71.23 & 57.29 & 2.23 & 33.92 & 41.29 & 49.46 & 52.53 & 29.22 & 39.73 \\
Unseen & 4.45  & 17.25 & 16.77 & 0.15 & 6.70  & 14.48 & 5.46  & 19.76 & 28.50 & 64.21 \\
Mean   & 7.58  & 27.77 & 25.94 & 0.28 & 11.19 & 21.44 & 9.84  & 28.72 & 28.86 & \textbf{49.09} \\
\midrule
\multicolumn{11}{c}{\textit{Run 3 (Seed 456)}} \\
\midrule
Seen   & 23.56 & 73.63 & 57.94 & 2.58 & 34.20 & 42.76 & 54.86 & 52.53 & 29.22 & 48.33 \\
Unseen & 4.75  & 14.73 & 15.69 & 0.00 & 7.50  & 14.38 & 13.52 & 14.97 & 33.41 & 65.85 \\
Mean   & 7.91  & 24.55 & 24.69 & 0.00 & 12.30 & 21.52 & 21.70 & 23.30 & 31.18 & \textbf{55.74} \\
\midrule
\multicolumn{11}{c}{\textit{Run 4 (Seed 789)}} \\
\midrule
Seen   & 21.46 & 74.89 & 57.29 & 1.35 & 34.27 & 41.17 & 49.15 & 53.23 & 37.69 & 40.35 \\
Unseen & 2.81  & 12.46 & 14.73 & 0.61 & 9.50  & 14.63 & 6.83  & 17.25 & 20.84 & 54.37 \\
Mean   & 4.97  & 21.36 & 23.44 & 0.84 & 14.88 & 21.59 & 11.99 & 26.05 & 26.84 & \textbf{46.32} \\
\midrule
\multicolumn{11}{c}{\textit{Run 5 (Seed 2025)}} \\
\midrule
Seen   & 22.86 & 74.19 & 56.84 & 1.99 & 34.43 & 42.18 & 57.59 & 52.33 & 31.93 & 38.09 \\
Unseen & 5.96  & 12.46 & 8.98  & 0.91 & 8.50  & 14.48 & 5.60  & 20.72 & 27.78 & 66.12 \\
Mean   & 9.46  & 21.33 & 15.51 & 1.25 & 13.63 & 21.55 & 10.21 & 29.68 & 29.71 & \textbf{48.33} \\
\midrule
\multicolumn{11}{c}{\textit{Average $\pm$ Std}} \\
\midrule
Seen   & $23.33_{\pm 1.42}$ & $73.01_{\pm 1.55}$ & $57.55_{\pm 0.61}$ & $1.92_{\pm 0.47}$  & $34.29_{\pm 0.26}$ & $42.23_{\pm 1.05}$ & $52.98_{\pm 3.63}$ & $52.74_{\pm 0.39}$ & $32.06_{\pm 3.09}$ & $\mathbf{42.56_{\pm 4.48}}$ \\
Unseen & $5.07_{\pm 1.71}$  & $13.84_{\pm 1.92}$ & $13.34_{\pm 3.39}$ & $0.33_{\pm 0.36}$  & $7.80_{\pm 1.19}$  & $14.51_{\pm 0.09}$ & $7.13_{\pm 3.69}$ & $18.35_{\pm 2.28}$ & $31.02_{\pm 7.86}$ & $\mathbf{61.69_{\pm 5.27}}$ \\
Mean   & $8.22_{\pm 2.32}$  & $23.21_{\pm 2.62}$ & $21.49_{\pm 4.55}$ & $0.47_{\pm 0.49}$  & $12.67_{\pm 1.57}$ & $21.59_{\pm 0.16}$ & $12.32_{\pm 5.45}$ & $27.16_{\pm 2.53}$ & $30.80_{\pm 3.59}$ & $\mathbf{50.19_{\pm 3.61}}$ \\
\bottomrule
\end{tabular}}
\end{table}

\begin{table}[h]
\centering
\caption{GZSL results on AVASTCTU. Seen = Seen Acc (\%), Unseen = Unseen Acc (\%), Mean = H-Mean (\%). Best Mean in \textbf{bold}.}
\label{tab:gzsl_avastctu}
\resizebox{\textwidth}{!}{%
\begin{tabular}{l cccccccccc}
\toprule
\textbf{Class} & \textbf{CVAE-ZSL} & \textbf{MZSL} & \textbf{CLIP-Decoder} & \textbf{P2T} & \textbf{ZET-LLM} & \textbf{ProtoLLM} & \textbf{SMELL} & \textbf{FL-ZSL} & \textbf{TZSL} & \textbf{\methodname} \\
\midrule
\multicolumn{11}{c}{\textit{Run 1 (Seed 42)}} \\
\midrule
Seen   & 46.18 & 99.44 & 65.30 & 14.29 & 99.48 & 90.07 & 81.18 & 96.45 & 52.50 & 96.68 \\
Unseen & 22.33 & 14.90 & 35.73 & 0.33  & 0.00  & 31.40 & 25.40 & 30.60 & 58.62 & 42.07 \\
Mean   & 30.11 & 25.92 & 46.19 & 0.65  & 0.00  & 46.57 & 38.69 & 46.46 & 55.39 & \textbf{58.63} \\
\midrule
\multicolumn{11}{c}{\textit{Run 2 (Seed 123)}} \\
\midrule
Seen   & 40.81 & 99.44 & 62.03 & 14.29 & 99.48 & 90.48 & 81.13 & 95.61 & 18.50 & 88.28 \\
Unseen & 24.07 & 0.78  & 34.67 & 0.18  & 0.00  & 28.68 & 23.42 & 26.23 & 39.62 & 40.83 \\
Mean   & 30.28 & 1.55  & 44.48 & 0.36  & 0.00  & 43.56 & 36.34 & 41.17 & 25.22 & \textbf{55.84} \\
\midrule
\multicolumn{11}{c}{\textit{Run 3 (Seed 456)}} \\
\midrule
Seen   & 31.09 & 99.39 & 62.12 & 14.29 & 99.48 & 91.73 & 80.90 & 95.84 & 35.54 & 97.43 \\
Unseen & 13.92 & 11.35 & 19.78 & 0.23  & 0.00  & 27.55 & 23.17 & 29.72 & 11.62 & 39.87 \\
Mean   & 19.23 & 20.37 & 30.01 & 0.46  & 0.00  & 42.37 & 36.02 & 45.37 & 17.51 & \textbf{56.58} \\
\midrule
\multicolumn{11}{c}{\textit{Run 4 (Seed 789)}} \\
\midrule
Seen   & 37.53 & 99.44 & 63.80 & 14.29 & 99.48 & 89.74 & 81.18 & 96.40 & 83.65 & 95.89 \\
Unseen & 9.37  & 30.72 & 26.48 & 0.27  & 0.00  & 28.65 & 22.77 & 35.58 & 35.90 & 39.57 \\
Mean   & 14.99 & 46.94 & 37.43 & 0.52  & 0.00  & 43.43 & 35.56 & 51.98 & 50.24 & \textbf{56.02} \\
\midrule
\multicolumn{11}{c}{\textit{Run 5 (Seed 2025)}} \\
\midrule
Seen   & 29.79 & 99.67 & 37.27 & 14.29 & 99.48 & 90.30 & 81.13 & 95.14 & 0.00  & 97.43 \\
Unseen & 29.52 & 9.05  & 37.88 & 0.20  & 0.00  & 22.00 & 23.90 & 23.23 & 41.92 & 41.20 \\
Mean   & 29.65 & 16.59 & 37.58 & 0.39  & 0.00  & 35.38 & 36.92 & 37.35 & 0.00  & \textbf{57.91} \\
\midrule
\multicolumn{11}{c}{\textit{Average $\pm$ Std}} \\
\midrule
Seen   & $37.08_{\pm 6.82}$ & $\mathbf{99.48_{\pm 0.10}}$ & $58.10_{\pm 11.72}$ & $14.29_{\pm 0.00}$ & $99.48_{\pm 0.00}$ & $90.46_{\pm 0.76}$ & $81.10_{\pm 0.10}$ & $95.89_{\pm 0.55}$ & $38.04_{\pm 32.11}$ & $95.14_{\pm 3.89}$ \\
Unseen & $19.84_{\pm 8.10}$ & $13.36_{\pm 9.84}$ & $30.91_{\pm 7.57}$ & $0.24_{\pm 0.05}$  & $0.00_{\pm 0.00}$  & $27.66_{\pm 3.47}$ & $23.73_{\pm 0.91}$ & $29.07_{\pm 4.67}$ & $37.53_{\pm 16.90}$ & $\mathbf{40.71_{\pm 1.01}}$ \\
Mean   & $24.85_{\pm 7.23}$ & $22.27_{\pm 14.74}$ & $39.14_{\pm 6.46}$ & $0.48_{\pm 0.10}$  & $0.01_{\pm 0.01}$  & $42.26_{\pm 4.15}$ & $36.71_{\pm 1.09}$ & $44.46_{\pm 5.54}$ & $29.67_{\pm 23.09}$ & $\mathbf{57.00_{\pm 1.22}}$ \\
\bottomrule
\end{tabular}}
\end{table}

\begin{table}[h]
\centering
\caption{GZSL results on GOODREADS. Seen = Seen Acc (\%), Unseen = Unseen Acc (\%), Mean = H-Mean (\%). Best Mean in \textbf{bold}.}
\label{tab:gzsl_goodreads}
\resizebox{\textwidth}{!}{%
\begin{tabular}{l cccccccccc}
\toprule
\textbf{Class} & \textbf{CVAE-ZSL} & \textbf{MZSL} & \textbf{CLIP-Decoder} & \textbf{P2T} & \textbf{ZET-LLM} & \textbf{ProtoLLM} & \textbf{SMELL} & \textbf{FL-ZSL} & \textbf{TZSL} & \textbf{\methodname} \\
\midrule
\multicolumn{11}{c}{\textit{Run 1 (Seed 42)}} \\
\midrule
Seen   & 6.83  & 25.28 & 12.50 & 17.89 & 14.56 & 23.92 & 21.28 & 34.53 & 16.67 & 34.03 \\
Unseen & 36.53 & 18.72 & 31.00 & 4.58  & 31.28 & 24.75 & 36.18 & 17.65 & 0.00  & 47.50 \\
Mean   & 11.51 & 21.51 & 17.82 & 7.30  & 19.87 & 24.33 & 26.80 & 23.36 & 0.00  & \textbf{39.65} \\
\midrule
\multicolumn{11}{c}{\textit{Run 2 (Seed 123)}} \\
\midrule
Seen   & 7.72  & 26.89 & 13.06 & 26.69 & 15.81 & 24.27 & 21.64 & 32.75 & 16.67 & 25.06 \\
Unseen & 45.02 & 20.03 & 34.58 & 4.00  & 32.50 & 24.77 & 22.33 & 16.52 & 0.00  & 50.77 \\
Mean   & 13.18 & 22.96 & 18.96 & 6.96  & 21.27 & 24.52 & 21.98 & 21.96 & 0.00  & \textbf{33.55} \\
\midrule
\multicolumn{11}{c}{\textit{Run 3 (Seed 456)}} \\
\midrule
Seen   & 6.58  & 16.75 & 14.19 & 29.67 & 15.36 & 23.36 & 22.17 & 34.36 & 16.67 & 31.56 \\
Unseen & 42.53 & 31.13 & 33.23 & 4.40  & 29.18 & 24.78 & 25.93 & 15.30 & 0.00  & 43.73 \\
Mean   & 11.40 & 21.78 & 19.89 & 7.66  & 20.13 & 24.05 & 23.90 & 21.17 & 0.00  & \textbf{36.66} \\
\midrule
\multicolumn{11}{c}{\textit{Run 4 (Seed 789)}} \\
\midrule
Seen   & 8.67  & 27.11 & 13.75 & 28.53 & 15.50 & 24.45 & 20.81 & 36.33 & 16.67 & 31.47 \\
Unseen & 34.92 & 20.53 & 32.05 & 4.33  & 29.78 & 24.88 & 28.20 & 16.05 & 0.00  & 35.80 \\
Mean   & 13.89 & 23.37 & 19.24 & 7.52  & 20.39 & 24.67 & 23.94 & 22.26 & 0.00  & \textbf{33.50} \\
\midrule
\multicolumn{11}{c}{\textit{Run 5 (Seed 2025)}} \\
\midrule
Seen   & 5.33  & 15.56 & 12.86 & 21.97 & 15.06 & 23.54 & 20.92 & 31.78 & 16.67 & 30.67 \\
Unseen & 42.93 & 43.35 & 29.33 & 6.15  & 33.45 & 24.82 & 29.32 & 15.37 & 0.00  & 44.30 \\
Mean   & 9.49  & 22.90 & 17.88 & 9.61  & 20.77 & 24.16 & 24.41 & 20.72 & 0.00  & \textbf{36.24} \\
\midrule
\multicolumn{11}{c}{\textit{Average $\pm$ Std}} \\
\midrule
Seen   & $7.03_{\pm 1.25}$  & $22.32_{\pm 5.09}$ & $13.27_{\pm 0.69}$ & $24.95_{\pm 4.40}$ & $15.26_{\pm 0.48}$ & $23.91_{\pm 0.47}$ & $21.36_{\pm 0.50}$ & $33.95_{\pm 1.76}$ & $16.67_{\pm 0.00}$ & $\mathbf{30.56_{\pm 3.32}}$ \\
Unseen & $40.39_{\pm 4.40}$ & $26.75_{\pm 9.41}$ & $32.04_{\pm 2.02}$ & $4.69_{\pm 0.75}$  & $31.24_{\pm 1.79}$ & $24.80_{\pm 0.05}$ & $28.39_{\pm 4.57}$ & $16.18_{\pm 0.97}$ & $0.00_{\pm 0.00}$  & $\mathbf{44.42_{\pm 5.59}}$ \\
Mean   & $11.89_{\pm 1.72}$ & $22.50_{\pm 0.72}$ & $18.76_{\pm 0.90}$ & $7.81_{\pm 0.93}$  & $20.48_{\pm 0.55}$ & $24.34_{\pm 0.25}$ & $24.21_{\pm 1.54}$ & $21.89_{\pm 1.02}$ & $0.00_{\pm 0.00}$  & $\mathbf{35.92_{\pm 2.50}}$ \\
\bottomrule
\end{tabular}}
\end{table}

\begin{table}[h]
\centering
\caption{GZSL results on PETFINDER. Seen = Seen Acc (\%), Unseen = Unseen Acc (\%), Mean = H-Mean (\%). Best Mean in \textbf{bold}.}
\label{tab:gzsl_petfinder}
\resizebox{\textwidth}{!}{%
\begin{tabular}{l cccccccccc}
\toprule
\textbf{Class} & \textbf{CVAE-ZSL} & \textbf{MZSL} & \textbf{CLIP-Decoder} & \textbf{P2T} & \textbf{ZET-LLM} & \textbf{ProtoLLM} & \textbf{SMELL} & \textbf{FL-ZSL} & \textbf{TZSL} & \textbf{\methodname} \\
\midrule
\multicolumn{11}{c}{\textit{Run 1 (Seed 42)}} \\
\midrule
Seen   & 1.95  & 36.95 & 32.52 & 28.20 & 25.40 & 27.94 & 53.71 & 44.30 & 3.91  & 29.01 \\
Unseen & 48.47 & 27.87 & 36.24 & 1.64  & 19.85 & 25.72 & 1.90  & 24.19 & 25.08 & 39.33 \\
Mean   & 3.76  & 31.77 & \textbf{34.28} & 3.09  & 22.28 & 26.78 & 3.67  & 31.29 & 6.76  & 33.39 \\
\midrule
\multicolumn{11}{c}{\textit{Run 2 (Seed 123)}} \\
\midrule
Seen   & 0.05  & 27.15 & 31.79 & 31.87 & 27.05 & 24.15 & 51.13 & 44.24 & 3.71  & 30.79 \\
Unseen & 49.44 & 34.05 & 31.47 & 0.78  & 21.45 & 25.66 & 1.71  & 24.51 & 17.12 & 38.59 \\
Mean   & 0.11  & 30.21 & 31.63 & 1.52  & 23.93 & 24.88 & 3.30  & 31.54 & 6.10  & \textbf{34.26} \\
\midrule
\multicolumn{11}{c}{\textit{Run 3 (Seed 456)}} \\
\midrule
Seen   & 0.00  & 29.74 & 31.52 & 38.23 & 26.30 & 27.41 & 52.72 & 44.50 & 4.04  & 27.48 \\
Unseen & 49.48 & 29.99 & 30.74 & 1.72  & 19.37 & 25.53 & 1.02  & 25.07 & 11.57 & 40.18 \\
Mean   & 0.00  & 29.86 & 31.13 & 3.29  & 22.31 & 26.44 & 1.99  & 32.07 & 5.99  & \textbf{32.64} \\
\midrule
\multicolumn{11}{c}{\textit{Run 4 (Seed 789)}} \\
\midrule
Seen   & 1.78  & 34.97 & 35.43 & 30.65 & 27.15 & 25.50 & 50.93 & 44.24 & 3.38  & 28.15 \\
Unseen & 49.86 & 30.56 & 30.08 & 1.43  & 20.85 & 25.63 & 1.81  & 23.92 & 19.87 & 40.89 \\
Mean   & 3.43  & 32.62 & 32.53 & 2.73  & 23.59 & 25.57 & 3.50  & 31.05 & 5.77  & \textbf{33.34} \\
\midrule
\multicolumn{11}{c}{\textit{Run 5 (Seed 2025)}} \\
\midrule
Seen   & 0.00  & 25.23 & 32.85 & 34.18 & 25.78 & 23.98 & 52.98 & 43.58 & 4.30  & 29.27 \\
Unseen & 50.27 & 38.02 & 30.53 & 1.01  & 21.75 & 25.59 & 1.69  & 24.14 & 10.03 & 38.24 \\
Mean   & 0.00  & 30.33 & 31.65 & 1.95  & 23.60 & 24.76 & 3.28  & 31.07 & 6.02  & \textbf{33.16} \\
\midrule
\multicolumn{11}{c}{\textit{Average $\pm$ Std}} \\
\midrule
Seen   & $0.76_{\pm 1.01}$  & $30.81_{\pm 4.49}$ & $32.82_{\pm 1.55}$ & $32.63_{\pm 3.40}$ & $26.34_{\pm 0.77}$ & $25.80_{\pm 1.82}$ & $52.29_{\pm 1.08}$ & $44.17_{\pm 0.35}$ & $3.87_{\pm 0.35}$  & $28.94_{\pm 1.26}$ \\
Unseen & $49.51_{\pm 0.67}$ & $32.10_{\pm 3.57}$ & $31.81_{\pm 2.53}$ & $1.32_{\pm 0.36}$  & $20.65_{\pm 1.02}$ & $25.63_{\pm 0.07}$ & $1.63_{\pm 0.31}$  & $24.37_{\pm 0.45}$ & $16.73_{\pm 6.15}$ & $\mathbf{39.45_{\pm 1.10}}$ \\
Mean   & $1.46_{\pm 1.95}$  & $30.96_{\pm 1.05}$ & $32.24_{\pm 1.25}$ & $2.52_{\pm 0.68}$  & $23.14_{\pm 0.78}$ & $25.69_{\pm 0.91}$ & $3.15_{\pm 0.59}$  & $31.41_{\pm 0.42}$ & $6.13_{\pm 0.37}$  & $\mathbf{33.36_{\pm 0.58}}$ \\
\bottomrule
\end{tabular}}
\end{table}

\begin{table}[h]
\centering
\caption{\methodname performance under varying numbers of labeled samples per unseen family ($K$-shot) across cybersecurity datasets. $K=0$ is the zero-shot setting; $S$ = Seen Acc (\%), $U$ = Unseen Acc (\%), $H$ = Harmonic Mean (\%). }
\label{tab:kshot}
\resizebox{0.6\textwidth}{!}{%
\begin{tabular}{c ccc ccc ccc}
\toprule
& \multicolumn{3}{c}{\textbf{CIC-AndMAL-2020}}
& \multicolumn{3}{c}{\textbf{BODMAS}}
& \multicolumn{3}{c}{\textbf{AVASTCTU}} \\
\cmidrule(lr){2-4}\cmidrule(lr){5-7}\cmidrule(lr){8-10}
\textbf{$K$}
    & $S$   & $U$   & $H$
    & $S$   & $U$   & $H$
    & $S$   & $U$   & $H$ \\
\midrule
\rowcolor{lightgray}
0 \small{(zero-shot)}
    & 66.56 & 51.07 & 57.78
    & 77.75 & 47.67 & 59.10
    & 99.58 & 34.28 & 51.01 \\
1
    & 57.41 & 64.45 & 60.73
    & 80.69 & 55.28 & 65.61
    & 95.80 & 54.96 & 69.85 \\
\rowcolor{lightgray}
2
    & 61.53 & 61.36 & \textbf{61.44}
    & 78.94 & 66.50 & \textbf{72.19}
    & 95.75 & 63.76 & 76.55 \\
5
    & 57.53 & 61.88 & 59.62
    & 80.69 & 65.13 & 72.08
    & 95.84 & 68.17 & \textbf{79.67} \\
\bottomrule
\end{tabular}}
\end{table}

\end{document}